# A novel approach to nose-tip and eye corners detection using H-K Curvature Analysis in case of 3D images

Parama Bagchi
Dept of Computer Science & Engineering
MCKV Institute of Engineering,
Kolkata 711204, India
paramabagchi@gmail.com

Debotosh Bhattacharjee
Dept of Computer Science & Engineering
Jadavpur University
Kolkata-700032, India
debotosh@ieee.org

Mita Nasipuri
Department of Computer Science & Engineering
Jadavpur University
Kolkata-700032, India
mitanasipuri@gmail.com

Dipak Kumar Basu
Department of Computer Science & Engineering
Jadavpur    University
Kolkata-700032,   India
dipakkbasu@gmail.com

*Abstract* — **In this paper we present a novel method that combines a HK curvature-based approach for three-dimensional (3D) face detection in different poses (X-axis, Y-axis and Z-axis). Salient face features, such as the eyes and nose, are detected through an analysis of the curvature of the entire facial surface. All the experiments have been performed on the FRAV3D Database. After applying the proposed algorithm to the 3D facial surface we have obtained considerably good results i.e. on 752 3D face images our method detected the eye corners for 543 face images, thus giving a 72.20% of eye corners detection and 743 face images for nose-tip detection thus giving a 98.80% of good nose tip localization**

## I. INTRODUCTION

Face detection and recognition have been the areas of key interest over the years, particularly feature extraction. In this paper, we propose an innovative approach to feature detection i.e. eye corners and nose-tip detection which can be defined as follows: "Given an arbitrary 3D face image in any pose (i.e. along X-axis, Y-axis and Z-axis), the goal of our algorithm is to detect the features i.e. to determine nose-tip and eye-corners and if present, return the coordinates of the located feature". In this paper, we propose a 2.5D feature extraction technique based on facial curvature analysis which is able to correctly localize the facial points. Most of the literature concerned face detection investigates face detection in two-dimensional (2D) images. In the majority of publications concerning 3D face recognition, faces are manually detected and registered in a standard position (sometimes referred to as "normal position") using some interactive tools that allow the selection of predefined landmark points, such as the position of the eyes. In some cases face localization is automated on the basis of strong assumptions about position and orientation with respect to the imaging device, such as assuming that the subject's nose is the point nearest to the camera [2, 3 ]. It is worth to mention that the problem of detecting faces in 3D acquisitions has yet to be thoroughly explored. In this paper, we propose an innovative 3D approach: our method assumes that the faces are rotated across any pose variations with respect to the X, Y and Z axes, the only restriction being that no self-occlusions and/or face hide the central part of the face containing the eyes and the nose of the subjects in the acquired scene. In Section II, some related works have been discussed. In Section III, a description of our method is enlisted. In Section IV, experimental results are discussed and finally in the Section V, the conclusions are enlisted. H-K classification is a method for curvature analysis on a 3D facial surface. H stands for Mean curvature and K stands for Gaussian curvature. Using the curvature concept, significant features like nose tips and eye- corners are detected. Based on a pre-defined HK classification table, an analysis of the 3D surface is done based on the signs of the components of the HK table.

## II. RELATED WORK

In this section we are going to discuss some related work done in the field of 3D curvature based face recognition. Chen[4] used a method based on mean and Gaussian curvatures. While HK-classification is rotation invariant, the way they analyze it with the assumption that the face model is in a frontal position to localize the nose tip and eye inner corners, makes their technique sensitive to roll and yaw rotation. HK-classification was also used by Colombo[5] for locating the same three main face points. However, their validation step is based on correlation with a template of a cropped face around the three main points, makes their solution very sensitive to yaw and pitch rotations. Moreover, a depth map used as template is very sensitive to spikes which may significantly change depth map appearance. Another work proposed by Sun [6], also relies on HK classification for automatic facial pose estimation. Pose is estimated by locating the three main points as well. They proposed to localize correct eye regions using statistical distances between the eyes and to remove clusters which have less number of points. Removing concave regions in such a way may simply lead to the removal of the eye regions. Other works are based on the shape index, for example: Colbry et al. [7] and Lu et al. [8]. They analyzed the shape index values to find feature points of the face. Nevertheless, they assume that the eyes are above the nose in the face scan. Most of the existing works embed some a priori knowledge on face configuration into the algorithm and make use of facial geometry-based analysis to localize geometrically salient feature points such as nose tip, eye corners, etc . A very similar work was proposed by Lu et al. Their approach is based on the simple assumption that the nose tip should have the highest z value in a frontal position. The face model is then rotated around y axis at a fixed step and the point with the largest value in the z-axis in each rotation is considered as the nose tip candidate. But in all other orientations (i.e. along x- axis,y-axis ) nose-tip may not have the highest z value. All these works can be categorized by their way of using curvature analysis, resulting in a rotation sensitive. Despite of all the reported results, their algorithm is not invariant to the roll rotation (along z-axis).In most cases, the subject under consideration is rotated and then their feature points are detected. Also, a considerable amount of work has been done in the field of curve-based methods [9]. This method deals with the curved based method by proposing a multi-scale feature extraction algorithm using a rotation and translation invariant local surface curvature measure known as the curvedness. It is a positive number that captures the amount of curvature surrounding a local region around a surface point. Different values of the curvedness of a point are calculated at multiple scales fitting a surface to its neighborhood of different sizes. A set of reliable salient feature points is formed by finding the extrema from the scale-space representation of the input surface. The curvedness approach was tested against a variety of 3D models with different noise levels. The shortcomings of the method based on curvedness were that no discussions about curvedness based on various poses were analyzed. In contrast to all the above methods, our method for HK_curvature based method, can detect the eye-corners as well as the nose-tip in case of any pose. It is actually insensitive to skin-color, beard, hair and other details. We have one restriction in this case and that is we have ignored any occlusions caused by a individual's hands, spectacles or similar obstacles. We have done all the experiments on the FRAV3D database[1] and taken into considerations all various poses (i.e. rotation along X-axis, Y-axis and Z-axis). We have detected the feature points on the 3D map itself without rotating or translating the face. Fig.1 shows our present method for feature detection. In this paper we have implemented only the first part of the figure i.e upto feature detection.

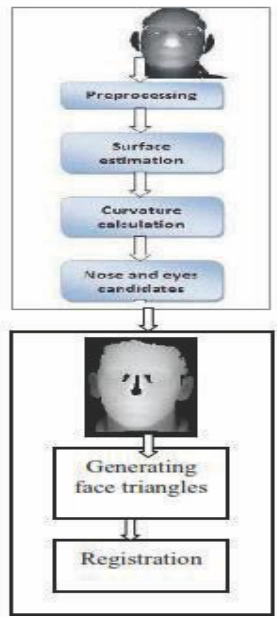

The present system for localization of nose-tip and eye-corners consists of the following four steps:
- Pre-processing
- Surface Estimation
- Curvature calculation
- Nose and eyes candidate generation

*A. Preprocessing :* First of all the 3D face of size [100 by 100] is cropped to size of dimensions[15 70].We are not considering the original image because presence of outliers may degrade the feature-detection rate. Next, the 3D image is thresholded by Otsu's thresholding algorithm. After thresholding the 3D images obtained are shown as in Figure 2.

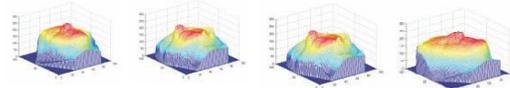

(a) Frontal pose (b) Rotated about y-axis    c) Rotated about axis
d) Rotated about z axis
Figure2. Mesh-grids after thresholding

Thresholding is basically done to remove irrelevant details like hair, beard and similar objects. In the next step the range image is smoothed by Gaussian filter. After smoothing the 3D images obtained are as shown in Fig 3.

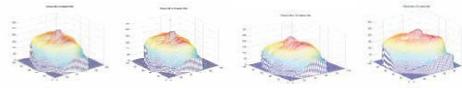

(a)Frontal pose (b) Rotated about y-axis   c) Rotated about axis  d) Rotated about z axis
Figure3. Mesh-grids after smoothing

*B. Surface Estimation:* A 3D surface is normally represented as a parametric equation:
$$y = f(x(t), y(t), z(t)) \quad \text{...............................(1)}$$

Every point on the 3D surface is made up of three coordinates $(x(t), y(t), z(t))$. After generating the 3D surface for each of the figures in Fig3(a),3(b),3(c),3(d) the corresponding surfaces as generated are as shown in Fig 4(a),4(b),4(c),4(d).

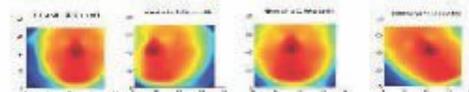

Figure 4. After smoothing, surface generation for images in (a) Frontal pose (b) Rotated about y-axis   c) Rotated about x-axis (d) Rotated about z-axis

*C. Curvature Estimation:* The present face detection method allows the faces, if any, to be freely oriented with respect to the camera plane, the only restriction being that no self-occlusions hide the eyes, the nose, or the area between them. The method analyzes a three-dimensional shot of a real scene given as a range image, that is an image where for each location (i, j) the coordinates are $(x_{ij}, y_{ij}, z_{ij})$. The curvature of a surface can be expressed in terms of first and second derivatives: let S be a surface defined by differentiable real valued function f :
$$S = \{(x, y, f(x, y))\} \quad (2)$$

The mean(H) and the Gaussian (K) curvature can be computed, at point $(x, y, f(x,y))$ can be calculated as follows[10]:

$$H(x, y) = \frac{(1+ f_y^2)f_{xx} - 2f_x f_y f_{xy} + (1+ f_x^2) f_{yy}}{2(1 + f_x^2 + f_y^2)^{3/2}} \quad (3)$$

$$K(x, y) = \frac{f_{xx} f_{yy} - f_{xy}^2}{(1 + f_x^2 + f_y^2)^2} \quad (4)$$

where $f_x$, $f_y$, $f_{xx}$, $f_{xy}$, $f_{yy}$ are the partial derivatives at (x,y). To estimate the partial derivatives, given the discrete representation prodsuced by the scanner, we consider a least squares biquadratic approximation for each element represented by the coordinates(i,j) in the range image:
$g_{ij}(x, y) = a_{ij} + b_{ij}(x - x_i) + c_{ij}(y - y_j) + d_{ij}(x - x_i)(y - y_j) + e_{ij}(x - x_i)^2 + f_{ij}(y - y_j)^2$ (5)
where the coefficients $a_{ij}$, $b_{ij}$, $c_{ij}$, $d_{ij}$, $e_{ij}$, $f_{ij}$ are obtained by the least squares fitting of the points in a neighborhood of $(x_i, y_j)$. Here $g_{ij}$ is the bi-quadratic approximation of the surface and $f_{ij}$ is the partial derivative of the surface. The derivatives of f in $(x_i, y_j)$ are then estimated by the derivatives of $g_{ij}$
$f_x(x_i, y_j) = b_{ij}$, $f_y(x_i, y_j) = c_{ij}$, $f_{xy}(x_i, y_j) = d_{ij}$, $f_{xx}(x_i, y_j) = 2e_{ij}$, $f_{yy}(x_i, y_j) = 2f_{ij}$.

The mean(H) and the Gaussian curvature(K) can then be estimated as follows:

$$H(x_{ij}, y_{ij}) = \frac{2(1 + c_{ij}^2)e_{ij} - b_{ij}c_{ij}d_{ij} + (1 + b_{ij}^2)f_{ij}}{(1 + b_{ij} + c_{ij})^{3/2}} \quad (6)$$

$$K(x_{ij}, y_{ij}) = \frac{4e_{ij}f_{ij} - d_{ij}^2}{(1 + b_{ij}^2 + c_{ij}^2)^2} \quad (7)$$

Analyzing the signs of the mean and the Gaussian curvature, we perform what is called an HK classification of the points of the region to obtain a concise description of the local behaviour of the region [11]. Table 1 shows the local behaviour of a 3D surface on the basis of H-K classification table [4].

TABLE 1. HK- CLASSIFICATION TABLE

| Values of H | Values of K | | |
|---|---|---|---|
| | K>0 | K=0 | K<0 |
| H<0 | Elliptical convex | Cylindrical Convex | Hyperbolic Convex |
| H=0 | Hyperbolic Symmetric | Planar | Impossible |
| H>0 | Elliptical concave | Cylindrical Concave | Hyperbolic Concave |

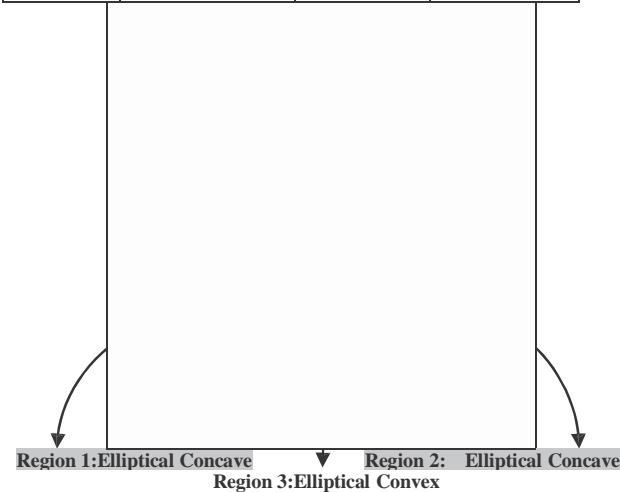

Region 1: Elliptical Concave    Region 2: Elliptical Concave
Region 3: Elliptical Convex

Figure 5. Segmented regions of a face from which facial features are extracted with the curvature map of region clearly classified.

For each region as shown in Fig 5 of adjacent points in the range images of the same HK class, the average mean and Gaussian curvatures are computed. In a smoothed face, those regions with the high three curvature values are the nose and the inside corners of the eyes. Three kind of points are obtained: darkest ones are hyperbolic points (K<0), intermediate grey level ones are elliptical convex points(H<0 and K >0), and brightest ones are elliptical concave points(H>0 and K>0). Cylindrical points (if K=0 and H<0 then the point is convex cylindrical; and if H>0, then it is concave cylindrical) and planar points (H=0 and K=0) do not appear since it would be difficult to obtain exactly zero curvature values. These regions can be clearly defined by applying suitable thresholds. In case of our algorithm eye corners and nose-tip were generated in the elliptical concave regions and convex regions respectively.

*D. Nose and eyes candidate generation:* After potential smoothing, the two different algorithms returns the eye-corners and the nose tips in case of any pose. For detecting the eye corners the elliptical concave region is thresholded by applying a threshold value of over 0.0001. For detecting nose-tip the elliptical convex region is thresholded by applying a threshold value of over 0.0001 then we have applied a maximum intensity technique to extract the nose-tip correctly.

**Algorithm 1 : Detect_ EyeCorners**
Input    : 3D Image
Output       : Eye-corners coordinates returned.
Step 1:  Generate H and K curvature map.
Step 2:  Initialize the value of p to 0
Step 3: Run loop for I from 1 to width of Image
Step 4: Run loop for J from 1 to height of Image
Step 5:  Check if H >0 and K>0.0001
Step 6:  arr[p]=K
Step 7:  p=p+1
Step 8: End if
Step9: End inner for loop
Step 10: End outer for loop
Step11: Sort the array in descending order.
Step12: Extract the two elements first and second Gaussian curvature values.

For detecting the nose-tip we first threshold Gaussian curvature above 0.0001. The highest curvature for Gaussian will come on the nose-tips. The algorithm for finding out the nose-tips is as follows:

**Algorithm 2 : Detect_ NoseTip**
Input    : 3D Image
Output    : Nose-tips value returned
Step 1: Generate H and K curvature map and generate intensity map for each point in the 3D image.
Step 2: Run loop for I from 1 to width of Image
Step 3: Run loop for J from 1 to height of Image
Step 4: Check if H<0 and K>0 for nose-tips
Step 5: Check for K>0.0001
Step 6:      arr[p]=K
Step 7:      p=p+1
Step 8:   End if
Step 7: End if
Step 8:   End inner for loop
Step 9:   End outer for

Step 10: Sort the array in descending order.
Step 11: Extract the first 5 points from the array
Step 12: - The point with maximum intensity is the nose-tip.

*Analysis of results for eye corners candidate generation*:

x *Frontal Pose*: In the Fig 6, we present an 3D image in frontal pose with the eye corners localized by HK classification method.

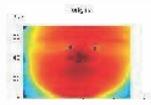

Figure 6. A sample output in frontal pose for detecting eye -corners

After running the algorithm the output obtained is as shown:

```
Row   Col   Curvature
 51   29    0.000410
 50   49    0.000225
```

The points of highest curvature values are the inner corners of the eye-region x *Non-frontal Pose*: -In the Fig 7 we present a 3D image in a non-frontal pose that is a 3D image rotated about Z axis with eye-corners marked in blue localized by HK.

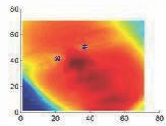

Figure 7. A sample output for an image in rotated pose(rotated about z-axis)for detecting eye –corners

After running the algorithm the output obtained is as shown:

```
Row   Col   Curvature
 51   37    0.000357
 43   18    0.000184
```

The points of highest curvature are the inner corners of the eye-region. Curvature measures give a distinct number of points surrounding eye-corners. But even if some points do come concentrating on other landmark points the max curvatures are on the eye-corners as denoted by the values

*Analysis of results for nose-tip candidate generation*:

x *Frontal Pose:* - In the Fig 8,we present a 3D image in frontal pose with the nose-tips localized by HK classification method.

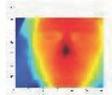

Figure 8 . A sample output in frontal pose for detecting nose-tip

After running the algorithm the output obtained is as shown:

```
Row   Col   Curvature   Intensity
 97   65    0.003453    161
 97   66    0.003108    161
  3   89    0.000892     64
 97   67    0.000819    159
 60   45    0.000563    254
```

Here, we first see that we have extracted the highest curvature values, and then the point with the maximum intensity which is eventually the nose-tip i.e. the points with coordinates (60, 45).

x *Non- Frontal Pose*: - In the Fig 9, we present an 3D image in frontal pose with the nose-tips localized by HK classification method.

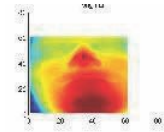

Figure 9. A sample output in rotated pose (rotated about x-axis) for detecting nose-tip

After running the algorithm the output obtained is as shown:

```
Row   Col   Curvature   Intensity
 50   53    0.092934    181
 51   51    0.00113     181
 50   53    0.00075     181
 26   41    0.00074     239
 44   39    0.00063     253
```

Here, we first see that we have extracted the highest curvature values, and then the point with the maximum intensity which is eventually the nose-tip(that is the points with coordinates (44,39).

## IV . EXPERIMENTAL RESULTS

In this section, we present a comparative analysis of how our proposed method outperforms the technique used in ref [12].The technique used in[12] used the maximum intensity technique to detect the nose-tip whose results are shown in Table 2.We have performed the test on 47 individuals including both male and female. First, we present the results obtained while detecting the nose-tip. The figure 10 shows some of the samples taken for nose-tip detection.

*A. Nose-tip detection results using Maximum Intensity Technique[12]:*

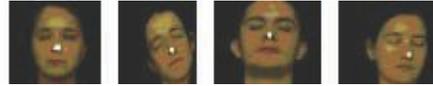

Figure10.Some samples from FRAV3D database with images in frontal pose, rotated about Z-axis, X-axis and y-axis.

TABLE 2. NOSE-TIP FOR FRONTAL AND NON-FRONTAL POSES IN CASE OF MAX INTENSITY TECHNIQUE

The notations for the column heads are given below:

A:Original Nose-tips in frontal pose
B: Nose-tips detected using maximum intensity technique

| 2 | Nose-tips detection in various poses | | | | | | | | | |
|---|---|---|---|---|---|---|---|---|---|---|
| A | B | C | D | E | F | G | H | I | J | K |
| 376 | 376 | +30 | 37 | 32 | +15 | 30 | 20 | +15 | 10 | 10 |
| | | -30 | 38 | 38 | -15 | 30 | 20 | -15 | 10 | 10 |
| | | +38 | 37 | 37 | +18 | 10 | 20 | +18 | 10 | 9 |
| | | -38 | 38 | 37 | -18 | 10 | 4 | -18 | 20 | 19 |
| | | +40 | 17 | 15 | +60 | 7 | 2 | +30 | 22 | 22 |
| | | -40 | 21 | 20 | -60 | 7 | 3 | -30 | 22 | 22 |

C: Angles of different viewpoint around Y-axis
D: Original Nose-tips in rotated pose around Y-axis
E: Nose-tips detected using maximum intensity technique
F: Angles of different viewpoint around X-axis
G: Original Nose-tips in rotated pose around X-axis
H: Nose-tips detected using maximum intensity technique
I: Angles of different viewpoint around Z-axis
J: Original Nose-tips in rotated pose around Z-axis

In the above case the nose-tip recognition rate was 95.21%.

*B. Eye-Corners detection results using Maximum Technique[12]:*
Using this technique there is a total failure of the maximum intensity technique.

*C. Nose-tip detection results using Curvature Analysis :* In the present proposed algorithm the recognition rate, was far

better than the max. intensity technique. The results are enlisted are as in Table 3:

TABLE 3. NOSE-TIP DETECTION FOR FRONTAL AND NON-FRONTAL POSES IN CASE OF CURVATURE ANALYSIS.

| 3 | | Nose-tips detection in various poses | | | | | | | | |
|---|---|---|---|---|---|---|---|---|---|---|
| | *A* | *B* | *C* | *D* | *E* | *F* | *G* | *H* | *I* | *J* | *K* |
| | 376 | 376 | +30 | 37 | 34 | +15 | 30 | 20 | +15 | 10 | 10 |
| | | | -30 | 38 | 38 | -15 | 30 | 23 | -15 | 10 | 10 |
| | | | +38 | 37 | 38 | +18 | 10 | 23 | +18 | 10 | 10 |
| | | | -38 | 38 | 38 | -18 | 10 | 10 | -18 | 20 | 20 |
| | | | +40 | 17 | 16 | +60 | 7 | 6 | +30 | 22 | 22 |
| | | | -40 | 21 | 21 | -60 | 7 | 6 | -30 | 22 | 22 |

The notations for the column heads are given below:
A: Original Nose-tips in frontal pose
B: Nose-tips detected using our algorithm
C: Angles of different viewpoint around Y-axis
D: Original Nose-tips in rotated pose around Y-axis
E: Nose-tips detected using our algorithm
F: Angles of different viewpoint around X-axis
G: Original Nose-tips in rotated pose around X-axis
H: Nose-tips detected using our algorithm
I: Angles of different viewpoint around Z-axis
J: Original Nose-tips in rotated pose around Z-axis
K: Nose-tips detected using our algorithm

As compared to the previous method, the present technique has a recognition rate of 98.80%, a better improvement over the maximum intensity technique.

*D. Eye-corners detection results using Curvature Analysis:* The figure 11 shows some of the samples from the FRAV3D database taken for eye-corner detection.

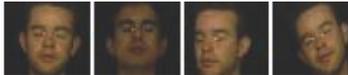

Figure11. Some samples from FRAV3D database with images in frontal pose, rotated about X-axis, Y-axis and Z-axis.

In the present proposed algorithm based on curvature analysis the results for eye-corner detection is enlisted in Table 4 :

TABLE 4. EYE-CORNERS DETECTION FOR FRONTAL AND NON-FRONTAL POSES USING CURVATURE ANALYSIS

| 4 | | Eye corners detection in various poses | | | | | | | | |
|---|---|---|---|---|---|---|---|---|---|---|
| | *A* | *B* | *C* | *D* | *E* | *F* | *G* | *H* | *I* | *J* | *K* |
| | 376 | 272 | +30 | 37 | 30 | +15 | 30 | 20 | +15 | 10 | 10 |
| | | | -30 | 38 | 30 | -15 | 30 | 20 | -15 | 10 | 10 |
| | | | +38 | 37 | 30 | +18 | 10 | 20 | +18 | 10 | 10 |
| | | | -38 | 38 | 30 | -18 | 10 | 10 | -18 | 20 | 10 |
| | | | +40 | 17 | 6 | +60 | 7 | 2 | +30 | 22 | 14 |
| | | | -40 | 21 | 6 | -60 | 7 | 3 | -30 | 22 | 10 |

The notations for the column heads are given below:
A: Original Eye-corners in frontal pose
B: Eye-corners detected using our algorithm
C: Angles of different viewpoint around Y-axis
D: Original Eye-corners in rotated pose around Y-axis
E: Eye-corners detected using our algorithm
F: Angles of different viewpoint around X-axis
G: Original Eye-corners in rotated pose around X-axis
H: Eye-corners detected using our algorithm
I: Angles of different viewpoint around Z-axis
J: Original Eye-corners in rotated pose around Z-axis
K: Eye-corners detected using our algorithm

As compared to the previous method that is the maximum intensity technique, the present technique has a recognition rate of 72.2%, a much better improvement over the maximum intensity technique which did not detect eye-corners at all. A comparative result of nose and eye-tips detection is shown in Fig 12 with blue bars representing success of eye-corners detection and red representing nose-tip detection.

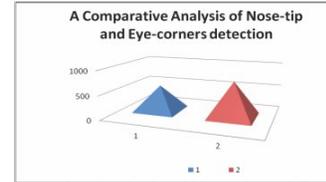

Figure12. A Comparative Analysis of Nose-tip and Eye-corners detection

V. CONCLUSION

In conclusion, it must be said that curvature analysis has indeed a great capability of detecting the eye-corners and the nose-tip of a 3D surface than the maximum intensity technique. Also we have proved that for large pose variations curvature analysis outperforms the intensity technique. Also the detection of eye-corners in case of curvature analysis whereby in max intensity technique, it is practically nil. Our future work aims at extracting the nose-tip and the eyecorners and thus detecting the region classifiers and finally registering the 3D images on the basis of these region classifiers by generating the face triangles which consists of the eye-corners and the nose-tip.